%% file: acra.tex
\documentclass{article}
\usepackage{acra}

\usepackage{algorithm}
\usepackage{algorithmic}
\usepackage{amsmath}
\usepackage{amssymb} 

\usepackage[pdftex]{graphicx}

\usepackage{url}

\newcommand{\E}{\mathbb{E}}

\newcommand{\lb}{\left [}
\newcommand{\rb}{\right ]}

\title{Training Directional Locomotion for Quadrupedal Low-Cost\\Robotic Systems via Deep Reinforcement Learning}
\author{Peter B{\"o}hm\thanks{Corresponding Author. The research for this paper received funding support from the Queensland Government through Trusted Autonomous Systems (TAS), a Defence Cooperative Research Centre funded through the Commonwealth Next Generation Technologies Fund and the Queensland Government}, Archie C. Chapman, and Pauline Pounds \\ The University of Queensland, Brisbane, Australia \\
	p.bohm@uq.edu.au, archie.chapman@uq.edu.au, pauline.pounds@uq.edu.au}

\begin{document}

\maketitle

\begin{abstract}
In this work we present Deep Reinforcement Learning (DRL) training of directional locomotion for low-cost quadrupedal robots in the real world. In particular, we exploit randomization of heading that the robot must follow to foster exploration of action-state transitions most useful for learning  both forward locomotion as well as course adjustments.
Changing the heading in episode resets to current yaw plus a random value drawn from a normal distribution yields policies able to follow complex trajectories involving frequent turns in both directions as well as long straight-line stretches.
By repeatedly changing the heading, this method keeps the robot moving within the training platform and thus reduces human involvement and need for manual resets during the training.
Real world experiments on a custom-built, low-cost quadruped demonstrate the efficacy of our method with the robot successfully navigating all validation tests. When trained with other approaches, the robot only succeeds in forward locomotion test and fails when turning is required.




\end{abstract}

\input{intro}

\input{related_work}

\input{prelims}

\input{method}

\input{experiments}

\input{results}

\input{conclusion}

\bibliography{bibliography}
\bibliographystyle{named}

\end{document}

%% file: intro.tex
\section{Introduction}
\label{sec:intro}

Legged systems can navigate uneven and irregular terrain that may be too challenging for wheeled or even tracked vehicles \cite{bledt2018cheetah}. This advantage comes at the cost of significantly more complex control, requiring intricate interplay of two or more legs, each comprising of multiple, separately actuated limbs, called gait. Changes in direction and/or velocity may require significant changes in gait.

\begin{figure}[t!]
	\centering
	\includegraphics[width=0.5\textwidth]{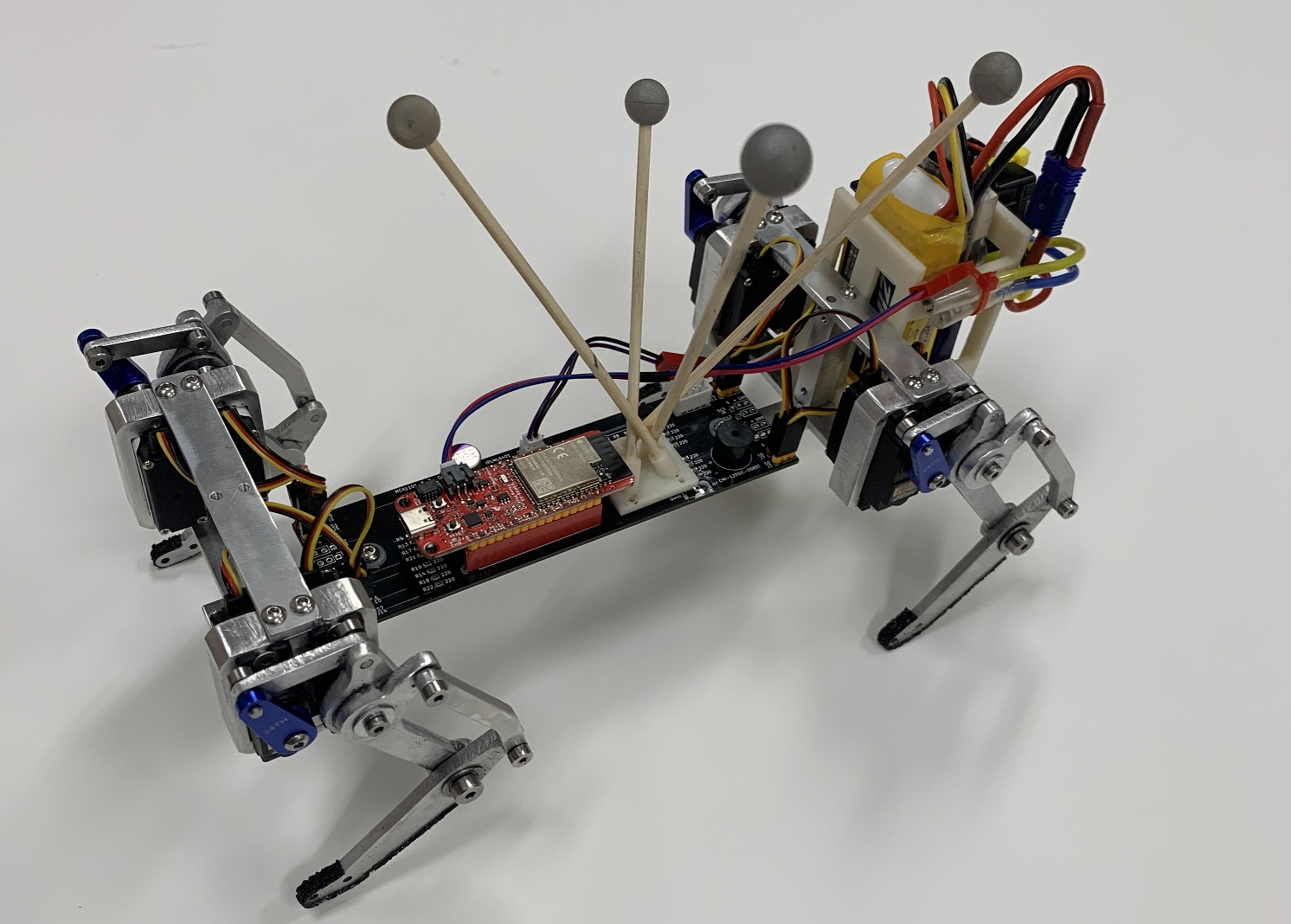}
	\vspace{-3.2mm}
	\caption{Quadrupedal robot platform.}
	\label{fig:quaid}
\end{figure}

One way of approaching the problem of quadrupedal locomotion is through application of classical control theory methods using state estimation, foot trajectory planning, and trajectory optimization.
They are often accomplished through Model-Predictive Control (MPC) and require accurate dynamics modeling as well as laborious manual tuning \cite{hutter2016anymal,bledt2018cheetah,sombolestan2021adaptive}.

A promissive alternative is model-free Deep Reinforcement Learning (DRL).
DRL has been making a steady progress from great successes in simulated and gaming domains \cite{silver2016mastering,mnih2013playing} into the real world. It has been successfully applied to various robot control tasks such as robotic arm manipulation \cite{liang2018gpu} and bipedal or quadrupedal locomotion \cite{xie2020learning,bloesch2021towards,lee2020learning,ha2021learning}.

Much of the DRL quadrupedal locomotion research is concentrated around a small number of highly capable, precise and high-cost robotic platforms, often complemented by a high fidelity simulation developed using highly accurate physics engines \cite{todorov2012mujoco,coumans2019} or custom actuation models \cite{lee2020learning}.

Despite the great engineering effort required to simulate these robots and their interaction with the surrounding world, there are still inevitable differences compared to the real-world, collectively known as the sim-to-real gap \cite{zhao2020sim}. This gap can be partially addressed through heavy use of domain randomization \cite{peng2018sim,xie2021dynamics}, trading optimality for robustness \cite{luo2017robust}. Randomizing various aspects of the environment, such as the latency, mass perturbations, proportional gain, and lateral push results in poor sample efficiency, longer training and requires more and faster computing resources.

Most current approaches do not control the actuators directly, but rather use the DRL network to generate control signals to PID/PD controllers which in turn actuate the hardware \cite{ha2021learning,wu2022daydreamer,smith2022walk}. In some cases, the DRL network is used to modulate the motion primitives via kinematic residuals. These high level commands are then processed through a foot trajectory generator and actuated through a high frequency joint PD controller \cite{lee2020learning}.
This approach relies on special high-cost hardware (e.g. the ANYmal robot), using bespoke actuators, as well as a custom built simulator, and a substantial on-board computing power.

This dependence on expensive specialty hardware and vast amounts of readily available compute precludes DRL trained robots from being practically used in the real world.
To use DRL at scale, it must support affordable commodity hardware that can be trained more efficiently.

Conversely, in our research, we focus on improving training efficiency when learning a directional locomotion (trajectory following task) for low-cost legged robots. 
We use a low-cost quadruped shown in Figure \ref{fig:quaid}, custom-built using commodity components. 
A DRL network provides the positional control signals directly to the servos without additional MPC or PID/PD controller in between. 
We train directly on the hardware because i) there is no simulation available, and ii) we want to test our approach in the noisy, low precision and non-stationary conditions emblematic of the real-world 

In this work we consider directional control of a low-cost DRL quadruped.
For a trained locomotion policy to be usable, it must not only move in a single direction shown in \cite{ha2021learning,wu2022daydreamer,smith2022walk}, but follow a desired trajectory.
In \cite{lee2020learning}, omnidirectional controller through curriculum based learning in simulation over millions of samples, however, such approach is not feasible when dealing with real-world robots.
Unlike in simulation, where the robot can be easily and arbitrarily reset at the start of each episode, the resets in real world require human intervention and there are more limits on which aspects can be randomized.

To learn the directional locomotion, the robot needs to essentially learn two components of the motion i) adjusting the heading, and ii) moving forward (which, due to the nature of quadrupedal walk may require frequent heading adjustments). Both components need to be balanced during the training.

Our approach is to set a new heading $\theta$ during episode resets, determined by adding a random value $\epsilon$ to the current yaw angle of the robot orientation. This $\epsilon$ is drawn from a normal distribution, which based on our experiments, provides a good balance of larger angles to train direction changes and smaller (or near zero) angles that allow the robot to move forward from the start of the episode. In the first case, the agent earns reward for minimizing the yaw angle from the heading, and in the second for distance traveled in the correct heading.

Trained using this approach, our robot is able to move forward, follow a circular trajectory as well as a complex figure eight trajectory. These changes in direction also keep the robot moving within the training platform and minimize human intervention when the robot moves out of the platform.






The structure of the paper is as follows:
Section~\ref{sec:related-work} discusses related work, followed by the technical preliminaries in
Section~\ref{sec:prelims}.
Section~\ref{sec:method} introduces the randomized heading episode resets. 
We describe the evaluation method in Section~\ref{sec:method}, and report the results in Section~\ref{sec:results}.
Section~\ref{sec:conclusion} concludes.

%% file: related_work.tex
\section{Related Work}
\label{sec:related-work}

Solving the problem of quadrupedal locomotion has long history in classic control theory and many of the well-known DRL robots come with a control theory based controller, usually utilizing model-predictive control (MPC).

\cite{hutter2016anymal} introduces ANYmal as a highly robust robot, specifically built for endurance, large mobility, and fast and dynamic locomotion skills. The accompanying classical ZMP planner is capable of walking, trotting, and stair climbing.

MIT's cheetah 3 is introduced in \cite{bledt2018cheetah}, a robust quadruped with expanded range of motions, higher forces and ability to control them in full 3D generality. The default MPC based controller provides three different gaits, including blind stair climbing.

There are multiple works demonstrating classic control theory approaches for Unitree A1 robot \cite{agrawal2022vision,sombolestan2021adaptive,ubellacker2021verifying}.

With custom-built simulators available for most of the high-end robots, many works investigate sim-to-real transfer. \cite{rudin2022learning} uses a massively parallel DRL to learn walking in minutes for ANYmal robot. 

\cite{tansim} improves the simulation fidelity by using a custom actuator model and including latency that results in successful sim-to-real transfer for Minitaur robot.

Applying multiple sources of perturbations and dynamics randomization targeted to specific parameters is done in \cite{xie2021dynamics} to enable sim-to-real transfer for Unitree Laikago.

Finally, there is a relatively small number of works exploring training directly on the hardware. 
\cite{ha2021learning} learns locomotion for Minitaur and focuses on automated learning that reduces human interventions by concurrently learning four policies: forward, backward, right, and left.

Real-world training on Unitree's A1 robot is investigated in \cite{wu2022daydreamer} and \cite{smith2022walk}, both approaches learn locomotion in a very short time, one hour and 20 minutes respectively. The second approach requires only 20k samples.


All previous DRL works utilize precise and robust high end hardware that has been shown to work using many other approaches.
In most of these works, only forward locomotion or one policy per direction is learned. For ANYmal, the omnidirectional policy is trained through simulated training over millions of samples. 
In contrast, we learn directional locomotion for a low-cost robot, custom-built using commodity components trained directly on the hardware.

%% file: prelims.tex
\section{Technical Preliminaries}
\label{sec:prelims}

\subsection{Reinforcement Learning}
Reinforcement learning (RL) is based on an abstraction of agent interacting with an unknown environment $\mathcal{E}$.
Based on the state observation $s \in \mathcal{S}$, the agent takes an action $a \in \mathcal{A}$ determined by the policy $\pi: \mathcal{S} \rightarrow \mathcal{A}$.
This action causes the environment to change, which is reflected by a new state observation $s'$ and the agent receives a reward $r \in \mathcal{R}^N$.
The goal is to learn a optimal policy $\pi$ that maximizes the expected discounted cumulative reward
$R_t = \sum_{i=t}^{T} \gamma^{i-t} r(s_i, a_i)$.
$\gamma$ is a discount factor used to set the priority of short-term rewards over long-term rewards.

For simple problems, optimal actions and estimates of their values can be learned using Q-learning, 
a form of temporal difference learning \cite{sutton1988learning}.
For most problems, however, the combination of state and action space is too large to learn the value of each action in each state.  
Instead of learning the entire Q-table, we can learn a parameterized value function $Q(s,a;\theta_t)$.
The standard Q-learning update for the parameters after taking action $A_t$ in state $S_t$ and observing the immediate reward $R_{t+1}$ and resulting state $S_{t+1}$ is then
$\theta_{t+1} = \theta_t + \alpha (Y^{\text{Q}}_t - Q(S_t,A_t;\theta_t)) \nabla_{\theta_t} Q(S_t,A_t;\theta_t)$
where $\alpha$ is a scalar step size and the target $Y^{\text{Q}}_t$ is defined as
$Y^{\text{Q}}_t \equiv R_{t+1} + \gamma \max_a Q(S_{t+1}, a; \theta_t) \,$.

In continuous action spaces, the goal is to optimize a parameterized policy $\pi_\phi$ by maximizing its expected return
	$J(\phi) = \E_{s_i \sim p_\pi, a_i \sim \pi} \lb R_0 \rb$.
One way of doing this is by \textit{deterministic policy gradient} introduced in \cite{silver2014deterministic}:
\begin{equation*} 
	\nabla_{\phi} J(\phi) = \E_{s \sim p_\pi} \lb \nabla_a Q^\pi(s,a)|_{a=\pi(s)} \nabla_{\phi} \pi_\phi(s) \rb. 
\end{equation*}
The DDPG algorithm \cite{ddpg}, is an extension of deterministic policy gradients through the use of deep neural networks.
The initial DDPG algorithm was further improved by \textit{twin delayed DDPG} (TD3) that addresses the over-estimation of Q-values in the critic network, specifically by innovating in three areas \cite{TD3}:
(i) using two critic networks, 
(ii) using delayed updates of the actor to reduce per-update error, and
(iii) applying action noise regularization.

\subsection{Dimensionality Reduction}
We modify the TD3 algorithm by encoding the sequence of observations through a \textit{Gated Recurrent Unit} (GRU), a type of recurrent neural network \cite{cho2014learning}. 
We use an untrained GRU network placed outside of the RL network, that serves as a variable sequence encoder that encodes the low-dimension representation of the state observation sequence before they are used for DRL training and inference. 

The sequence of observations from the start of the episode is thus reduced to a fixed length vector.
This vector is combined with the observations from the current step and the result is then stored in the replay buffer that is sampled during the training.
Observations are encoded cumulatively, with each step further unrolling the hidden state of the GRU.
We denote this algorithm R-TD3.

\subsection{Orientation Reset Strategies}
An important consideration for DRL training with real-world robots is the number of human interventions required. Unlike in simulation, where at the start of the episode, the robot can be positioned as needed, in the real-world, it is easier to manipulate the observations, rather than the physical robot.
We do this at the start of each episode by setting a new heading $\theta$, that the robot needs to follow.

The easiest way to do this to keep the heading constant throughout the training. Many simulated environments (Ant, Humanoid in OpenAI Gym, Minitaur in PyBullet Environments) use this strategy by only rewarding movement along the $x$ axis. During episode reset, the robot is repositioned to face the correct direction and some randomization applied to the the initial state. 

When dealing with physical robots, this approach has three main drawbacks: i) robot getting stuck facing the wrong direction for a number of episodes causing long periods of negative rewards, ii) as the policy improves, the robot will run out of the available space within one or two episodes requiring manual repositioning, and iii) the manual positioning tends to lead uneven distribution of the initial orientation with respect to the required heading which may lead to the trained policy being able to turn only in one direction. 

Another option is to change required heading during episode resets based on the current orientation of the robot. The new heading can be either set to the current orientation or randomized by adding a random angle $\epsilon$.

The first option makes it easier for the robot to move forward because it does not need to reposition itself first. Because the robot is not forced to turn, it may also lead to the trained policy handling turns poorly.

The second option requires the robot to adjust the heading at the start of each episode. The consideration here is how big this adjustment should be. Setting it too high may require too many steps to find the heading or it may lead to overly aggressive turning that overshoots and causes spinning behavior to be learned.

%% file: method.tex
\section{Apparatus and Method}
\label{sec:method}
To compare how different reset strategies affect the collected transitions and thus the policy training,
we train using three different heading reset strategies: setting $\theta$ to current yaw, random resets drawn from uniform distribution and random resets drawn from normal distribution.
We then validate the resulting policies on ability to follow three different trajectories to asses their relative performance.

\begin{figure}[t!]
	\centering
	\includegraphics[width=0.4\textwidth]{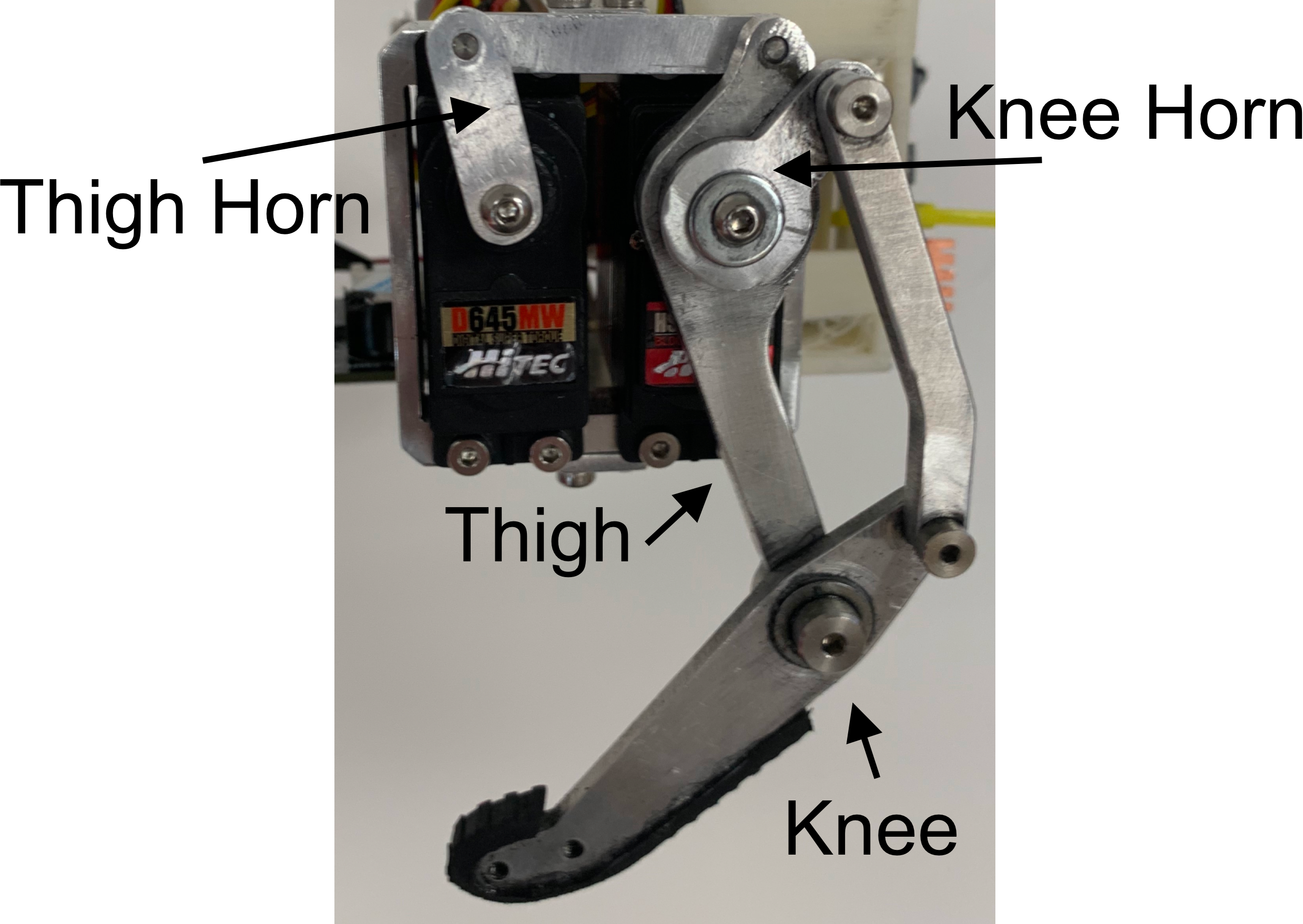}
	\vspace{-3.2mm}
	\caption{Leg consisting of thigh and knee links, actuated by commodity servos.}
	\label{fig:quaid_leg}
\end{figure}

\subsection{Robot Platform}
\label{sec:robot-platform}
We use a custom made low-cost quadrupedal robot shown in Figure~\ref{fig:quaid}. Each leg is actuated by commodity grade servos (Hitec D85MG), one controlling the thigh link and the second controlling the knee link, shown in Figure \ref{fig:quaid_leg}. The control signal is positional, providing a position trajectory for each servo.
The robot is controlled by an ESP32-S2 MCU. The communication between the MCU and the training machine is via WiFi and Pub/Sub protocol over a MQTT message broker.

\subsubsection{Observation Space}
The observation space is configurable, each component can be turned on and off using a configuration file. Observation space contains: i) time from the last action, ii) distance $\Delta{d}$ traveled in the set heading, iii) current drawn (only if above some threshold), iv) yaw angle from the set heading, v) $\Delta$yaw compared to the last time step, vi) roll, and vii) position of each servo.

Initially, the orientation and distance traveled were collected using an onboard IMU and time of flight sensor, however, due to the 2021 silicon shortage we were forced to use an OptiTrack tracking system. As such, the full observations are assembled from the OptiTrack and quadruped streamed independently.

\subsubsection{Action Space}
The action space is comprised of the position of each servo. Actions are streamed from the off-board workstation running the training and received by the onboard MCU. 
The actions are real numbers on the interval $[-1, 1]$ and these values are mapped to the allowed range of each servo. To prevent impermissibly large and abrupt movements, this value $u_t$ is passed through an exponential filter resulting in the actuation command $\bar{a}_t = \bar{a}_{t-1} * c + u_t * (1 - c)$. The filter coefficient $c$ is set in the configuration file, with an empirically determined value (we achieved best results with $c = 0.5$).

\subsubsection{Reward Function}
The weights in the reward function are configurable. We achieved the best results with the following: $R = \Delta{d} + \Delta{yaw} - 0.1 \times yaw - 0.1 \times roll - current$ rewards forward motion, yaw changes towards the set heading and marginally penalizes  differences from level orientation. 

\begin{figure}[t!]
	\centering
	\includegraphics[width=0.45\textwidth]{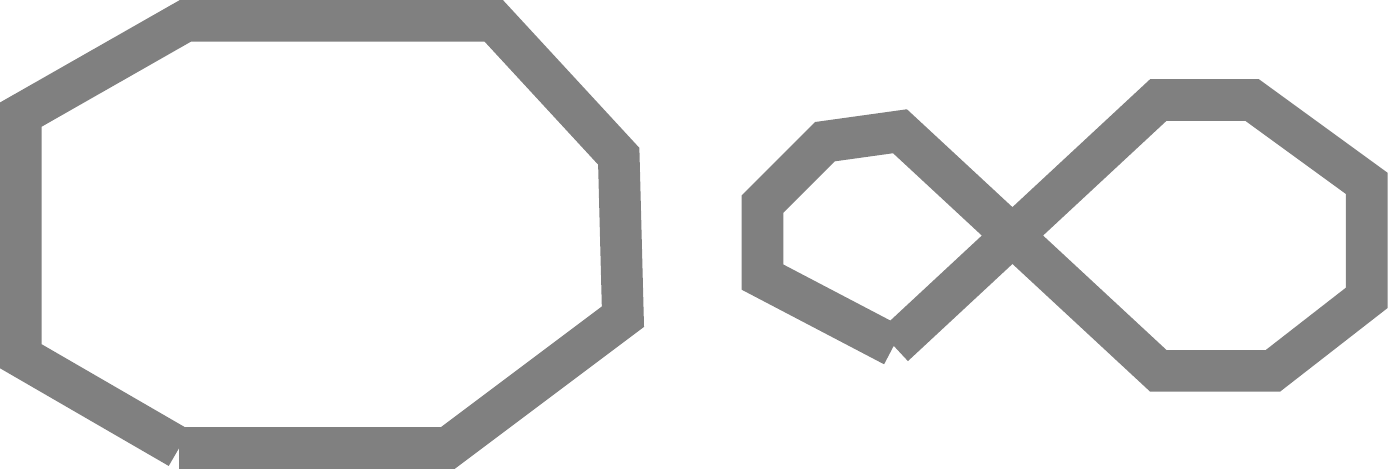}
	\vspace{-3.2mm}
	\caption{Quasi-circle and figure 8: trajectories used to validate the trained policies.}
	\label{fig:trajectories}
\end{figure}

\subsection{Training}
We run a standard DRL training process using the R-TD3 algorithm described in Section \ref{sec:prelims}. During training, the robot moves freely on the training platform ($3.5m \times 2.5m$ rectangle), with the goal of maximizing reward.

We train the robot using the following strategies to reset the heading between the episodes:

i) Setting $\theta$ to the current yaw so the robot does not need to adjust its heading, only move forward, 

ii) Setting $\theta$ to current yaw $+ \epsilon$, where $\epsilon$ is drawn from a uniform distribution $[-\pi/4, +\pi/4]$, and 

iii) Setting $\theta$ to current yaw $+ \epsilon$, where $\epsilon$ is drawn from a normal distribution with $\mu = 0$ and $\sigma = \pi/4$.

Each training run comprises 600 episodes of 500 steps, resulting in 300k steps per run. 
We then choose the best performing policy for each strategy, determined by the highest score achieved in non-exploration tests.
Selected policies are then validated by following predefined trajectories.

\subsection{Validation}
Trained policies are compared on ability to follow three trajectories, shown in Figure \ref{fig:trajectories}:

i) Straight line forward, 

ii) Quasi-circle, and 

iii) Figure eight. 

The trajectories are defined as a set of waypoints. 
A new heading $\theta$ is provided by an external controller every $5s$ based on the current position and the next waypoint using the following formula: $\theta = atan2(waypoint_y - position_y, waypoint_x - position_x)$.


%% file: experiments.tex

%% file: results.tex
\section{Results}
\label{sec:results}

\begin{figure*}[t!]
	\centering
	\includegraphics[width=1\textwidth]{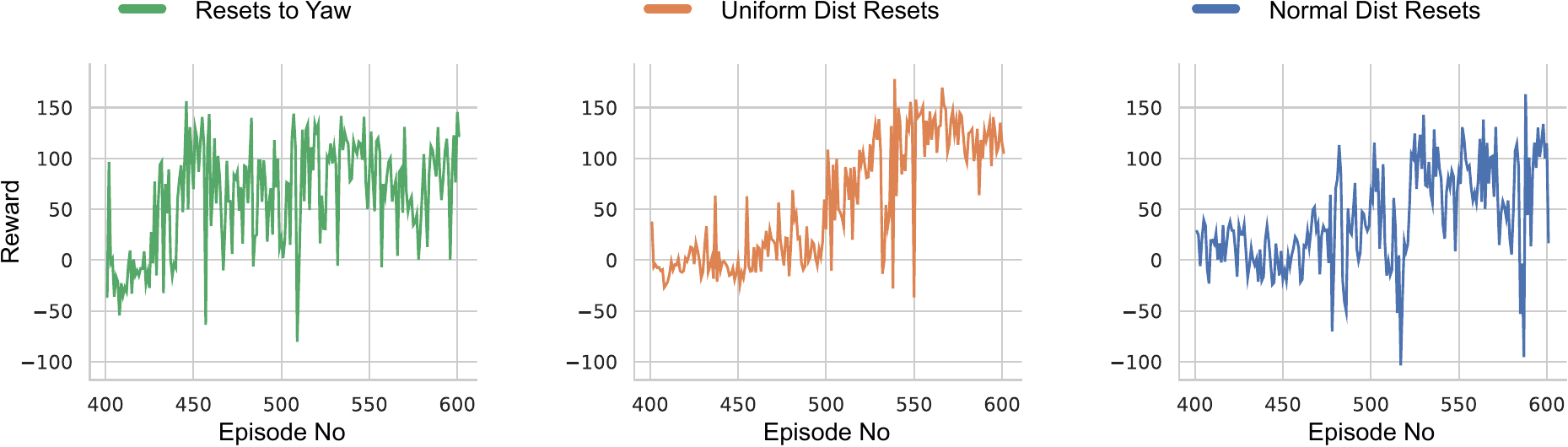}
	\vspace{-3.2mm}
	\caption{Learning curves using different strategies for heading resets. Showing the last 100k steps of training.}
	\label{fig:heading-resets-charts}
\end{figure*}

\subsection{Training Convergence Comparison}
Figure \ref{fig:heading-resets-charts} shows the learning curves for different strategies of heading resets (last 200 episodes / 100k steps). 

These results show that resetting the heading to yaw was fastest to converge; the robot did not need to adjust the direction at the start of the episode, and thus was able to quickly score points by moving forward. While the reward function rewards rotation towards the heading, a longer distance can be covered by moving forward, generating a larger reward.

The next to converge was the training using uniform distribution resets. This training also achieved the highest rewards. Finally, the slowest to converge was training using normal distribution resets. The policy struggled with larger $\epsilon$ values generated by the long tails of normal distribution.

\subsection{Validation Tests Comparison}
\begin{figure}[t!]
	\centering
	\includegraphics[width=0.5\textwidth]{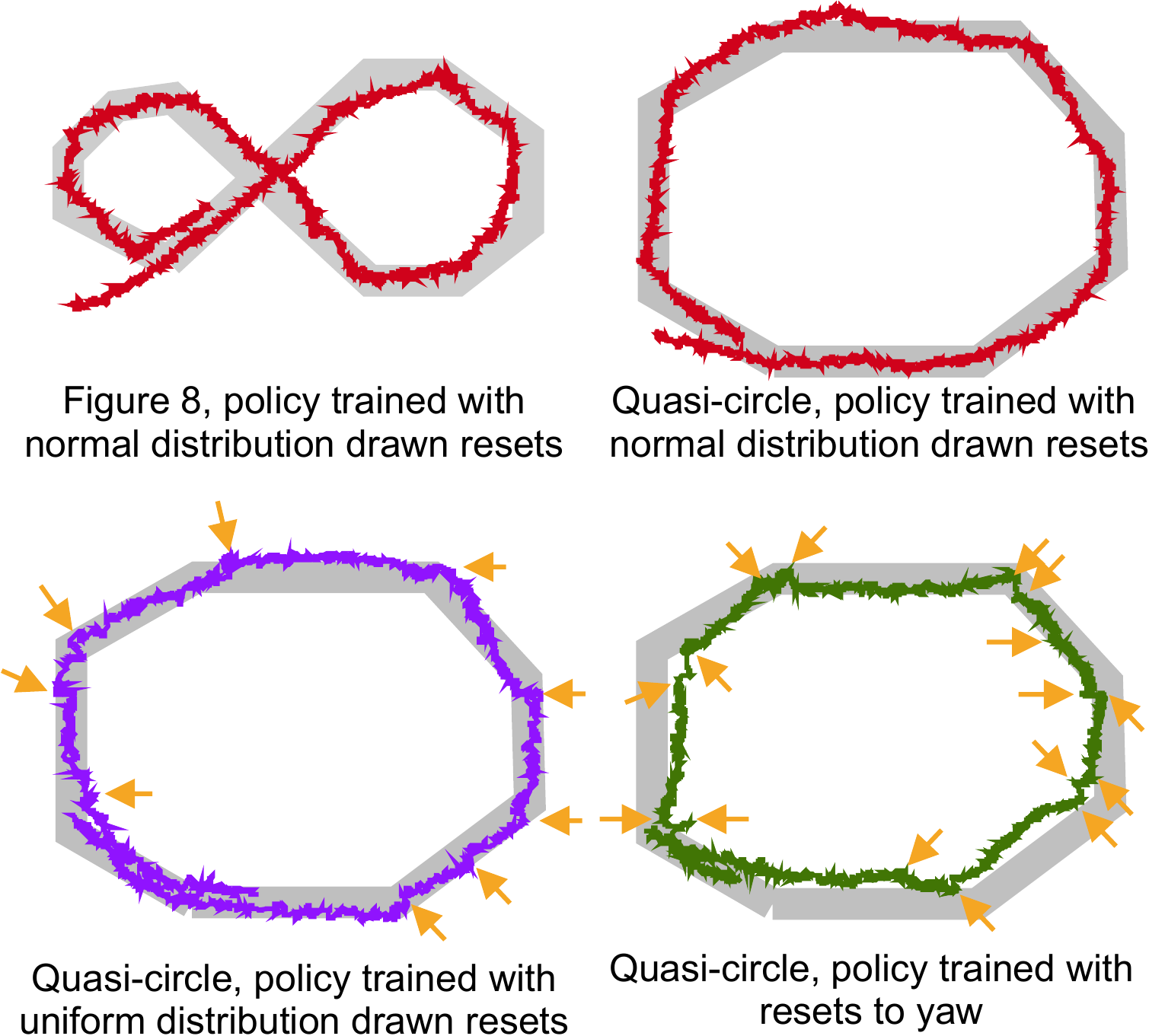}
	\vspace{-3.2mm}
	\caption{Trajectories recorded by running validation tests. The arrows show points where the robot stopped moving and required adjustment of heading. Neither uniform resets trained nor resets to yaw trained policy succeeded on the figure 8 course. }
	\label{fig:followed_trajectories}
\end{figure}

Trajectories recorded during the validation tests are shown in Figure~\ref{fig:followed_trajectories}.

\subsubsection{Straight Line Test}
All three policies were able to follow the straight line. The longest distance was covered by the policy trained with uniform resets.

\subsubsection{Quasi-circle Test}
This test requires turning in only small angles and only in one direction. Still the heading to yaw policy failed to complete the circle. When adjusting the heading, it overshot and was unable to deal with the large yaw/heading angle, rarely seen during the training. As such it required multiple interventions to decrease the heading angle.
Policy trained with uniform resets managed to complete the test, but required minor perturbations in several points where it struggle to adjust the heading.
Policy trained with normal distribution resets completed the test without any help.

\subsubsection{Figure Eight Test}
This is the most difficult test involving frequent turns in both directions as well as long straight-line stretches. Only the policy trained with random resets drawn from the normal distribution managed to complete this test. 

The policy with uniform resets handled the turns well, but only up to the interval of the distribution ($\pm\pi / 4$). Even slightly larger angles caused the robot to stop with the action commands stuck at the same values.
The robot following this trajectory is shown in the accompanying video.


\section{Discussion and Future Work}
Training directly on the hardware is especially useful for readily available and ubiquitous low-cost robotics systems for which a suitable, high-fidelity simulators are usually not available. The noisy and low-precision nature of these systems make them very difficult to model accurately.

The downside is that it often requires human interventions, and/or causes wear and tear to the hardware, which in turn creates a need for higher sample efficiency.
Our approach shows that even small changes to the environment setup can have large impact on the resulting trained policy. 
This is in line with recent findings \cite{rudin2022learning} in which simple modifications to the state and action spaces as well as the reward function lead to significantly shorter training times.  

The policy trained with normal distribution resets performs the best in the quazi-circle as well as the complex figure eight test. However, the policy trained with uniform rests perform the best in straight line test as it handles small course adjustments better than the other policies and it only fails on angles bigger than the distribution interval.

While the policy with resets to yaw spent the most time moving forward during the training, lack of transitions spent adjusting the heading left it unable to handle even small angle adjustments.

As such there could be advantages in combining multiple approaches, for example first training the policy using the uniform rests and once the policy is sufficiently trained, switch to normal distribution resets with larger standard deviation.

Another option is to use curriculum training and progressively increasing the uniform distribution interval. This could be combined with alternating between resets to yaw and randomized resets to generate transitions more equally distributed between the two components of directional locomotion.

%% file: conclusion.tex
\section{Conclusion}
\label{sec:conclusion}

This paper investigated using deep reinforcement learning for training directional quadrupedal locomotion directly on the hardware as a promissing alternative to DRL training in simulation coupled with dynamics randomization.
Unlike many other works, we focused on low-cost, low-precision robotics utilizing commodity components.
Through randomization of the robot heading $\theta$ during episode resets calculated as current yaw + $\epsilon$ drawn from a normal distribution with $\mu = 0$ and $\sigma = \pi/4$ during the training, the learned policy is able to follow a figure eight trajectory with multiple turns not seen during the training.